%% file: iclr2024_conference.tex
\documentclass{article} 
\usepackage{iclr2024_conference,times}

\input{math_commands.tex}

\usepackage{hyperref}
\usepackage{url}
\usepackage{graphicx} 
\usepackage{dsfont}
\usepackage{booktabs}
\usepackage{amsmath}
\usepackage{amssymb}
\usepackage{wrapfig} 
\usepackage{float}

\title{CAMIL: Context-Aware Multiple Instance Learning for Cancer Detection and Subtyping in Whole Slide Images}

\author{
    Olga Fourkioti\thanks{Corresponding author} \\
    The Institute of Cancer Research \\ London, United Kingdom \\
    \texttt{ofourkioti@icr.ac.uk} \\
    \And
    Matt De Vries \\
    The Institute of Cancer Research \\ London, United Kingdom \\
    \texttt{mvries@icr.ac.uk} \\
    \And
    Chen Jin \\
    Centre for Medical Image Computing \\
    Department of Computer Science \\
    University College London, United Kingdom \\
    \texttt{chen.jin@ucl.ac.uk} \\
    \And
    Daniel C. Alexander \\
    Centre for Medical Image Computing \\
    Department of Computer Science \\
    University College London, United Kingdom \\
    \texttt{d.alexander@ucl.ac.uk} \\
    \And
    Chris Bakal \\
    The Institute of Cancer Research \\ London, United Kingdom \\
    \texttt{cbakal@icr.ac.uk} \\
}

\iclrfinalcopy 
\begin{document}

\maketitle

\begin{abstract}
The visual examination of tissue biopsy sections is fundamental for cancer diagnosis, with pathologists analyzing sections at multiple magnifications to discern tumor cells and their subtypes. However, existing attention-based multiple instance learning (MIL) models used for analyzing Whole Slide Images (WSIs) in cancer diagnostics often overlook the contextual information of tumor and neighboring tiles, leading to misclassifications. To address this, we propose the Context-Aware Multiple Instance Learning (CAMIL) architecture. CAMIL incorporates neighbor-constrained attention to consider dependencies among tiles within a WSI and integrates contextual constraints as prior knowledge into the MIL model. We evaluated CAMIL on subtyping non-small cell lung cancer (TCGA-NSCLC) and detecting lymph node (CAMELYON16 and CAMELYON17) metastasis, achieving test AUCs of 97.5\%, 95.9\%, and 88.1\%, respectively, outperforming other state-of-the-art methods. Additionally, CAMIL enhances model interpretability by identifying regions of high diagnostic value\footnote{Our code is available at \url{https://github.com/olgarithmics/ICLR_CAMIL}.}.

\end{abstract}

\section{Introduction}

Deep learning (DL) methods have revolutionized the development of highly accurate diagnostic machines \citep{DBLP:journals/dsp/MoralesEN21} that rival or even surpass the performance of expert pathologists \citep{
DBLP:journals/mia/TongWGGHR14,
DBLP:journals/tmi/MelendezGMPRBAMAS15,
DBLP:journals/tmi/QuellecLCCC16,
DBLP:conf/isbi/DasCRCS18,
DBLP:journals/corr/abs-1912-12378,wang_predicting_2021}. These advancements have been facilitated by the emergence of weakly supervised learning, which eliminates the need for laborious pixel-level annotations. Models trained using weakly supervised learning, relying solely on slide-level labels, have demonstrated exceptional classification accuracy on whole slide imaging (WSI) data, paving the way for scalable computational decision support systems in clinical practice \citep{DBLP:journals/mia/XuZCLT14,DBLP:journals/corr/abs-1802-02212,DBLP:conf/iccv/XuSSKYLWMX19,DBLP:journals/cmig/ZhouJCHHWZCGL21}.

In the context of cancer histopathology, WSIs are not processed as a single image by DL models. Instead, WSIs are frequently subdivided into smaller tiles, which serve as an input. The task is, then, to classify the WSI based on the features extracted from the individual tiles. Most current methods for weakly supervised WSI classification use the Multiple Instance Learning (MIL) framework, which considers each WSI as a `bag' of tiles and attempts to learn the slide-level label without prior knowledge about the labels of the individual tiles.

A major bottleneck in the deployment of MIL models, and the weakly-supervised learning paradigm in general, is that the MIL model is either permutation invariant, meaning that the tiles within a WSI exhibit no ordering among each other \citep{DBLP:conf/midl/SharmaSEMSB21,
DBLP:conf/midl/XieMVCYCF20}, or permutation-aware without explicit information guidance. In other words, the spatial relationship of one tile to another is either ignored, or the dependencies between the tiles are implicitly modeled during training without requiring direct instructions \citep{DBLP:conf/nips/ShaoBCWZJZ21, 10.1093/bioinformatics/btaa847,
Campanella2019ClinicalgradeCP}.

However, explicit knowledge about a tile's spatial arrangement is particularly relevant in cancer histopathology, where cancer and normal cells are not necessarily distributed randomly inside an image. Contextual insights into the cellular landscape, such as the spatial dispersion of cells, the arrangement of cell clusters, and the broader characteristics of the tissue microenvironment, provide a more comprehensive view of the tile's local environment, enabling a better assessment of subtle variations and abnormalities that may indicate the presence of cancer.

In this paper, we propose a novel framework dubbed Context-Aware Multiple Instance Learning (CAMIL) to harness the dependencies among the individual tiles within a WSI and impose contextual constraints as prior knowledge on the multiple instance learning model. By explicitly accounting for contextual information, CAMIL aims to enhance the detection and classification of localized tumors and mitigate the potential misclassification of isolated or noisy instances, thereby contributing to an overall improvement in performance for both individual tiles and WSIs. Moreover, the attention weights enhance the interpretability of the model by highlighting sub-regions of high diagnostic value within the WSI. 

\section{Related work}\label{sec:related_work}

Under the MIL formulation, the prediction of a WSI label (i.e., cancerous or not) can come either directly from the tile predictions (instance-based) \citep{Campanella2019ClinicalgradeCP,10.1093/bioinformatics/btaa847,DBLP:conf/cvpr/HouSKGDS16,DBLP:conf/iccv/XuSSKYLWMX19}, or from a higher-level bag representation resulting from the aggregation of the tile features (bag embedding-based) \citep{DBLP:conf/icml/IlseTW18,DBLP:journals/corr/abs-2004-09666,DBLP:conf/midl/SharmaSEMSB21,DBLP:journals/pr/WangYTBL18}. The bag embedding-based approach has empirically demonstrated superior performance \citep{DBLP:conf/midl/SharmaSEMSB21,DBLP:journals/pr/WangYTBL18}. Most recent bag embedding-based approaches employ attention mechanisms \citep{DBLP:conf/nips/VaswaniSPUJGKP17}, which assign an attention score to every tile reflecting its relative contribution to the collective WSI-level representation. Attention scores enable the automatic localization of sub-regions of high diagnostic value in addition to informing the WSI-level label \citep{
DBLP:conf/cvpr/ZhangMAGSVS21,
DBLP:conf/miccai/BenTaiebH18,
DBLP:journals/corr/abs-2004-09666}. 

Attention-based MIL models vary in how they explore tissue structure in WSIs. Many are permutation invariant, assuming the tiles are independent and identically distributed. Building upon this assumption, \citet{DBLP:conf/icml/IlseTW18} proposed a learnable attention-based MIL pooling operator that computes the bag embedding as the average of all tile features in the WSI weighted by their respective attention score. This operator has been widely adopted and modified with the addition of a clustering layer \citep{DBLP:journals/corr/abs-2004-09666,DBLP:journals/cbm/LiLSYWSA21,DBLP:journals/mia/YaoZJHH20} to further encourage the learning of semantically-rich, separable and class-specific features. Another variation of the same model uses `pseudo bags' \citep{DBLP:conf/cvpr/ZhangMZQYCZ22}, splitting the WSI into several smaller bags to alleviate the issue of the limited number of training data. Recently, data augmentation has been adopted to inflate the number of bags \citep{10.1007/978-3-031-43987-2_46, liu2023pseudobag, shao2023augdiff}.

However, permutation invariant operators cannot inherently capture the structural dependencies among different tiles at the input. The lack of bio-topological information has partially been remedied by the introduction of feature similarity scores instead of positional encodings to model the mutual tile dependencies within a WSI \citep{DBLP:conf/midl/XieMVCYCF20,DBLP:journals/pami/TellezLLC21,DBLP:conf/cvpr/AdnanKT20}. For instance, DSMIL \citep{li2021dual} utilizes a non-local operator to compute an attention score for each tile by measuring the similarity of its feature representation against that of a critical tile. To consider the correlations between the different tiles of a WSI, transformer-based architectures have been introduced, which usually make use of a learnable position-dependent signal to incorporate the spatial information of the image \citep{DBLP:conf/cvpr/ZhaoYFLZZSYMFY20,DBLP:journals/corr/abs-1906-04881}. For instance, TransMIL \citep{DBLP:conf/nips/ShaoBCWZJZ21} is a transformer-like architecture trained end-to-end to optimize for the classification task and produce attention scores while simultaneously learning the positional embeddings. 

In CAMIL, we provide explicit guidance regarding the context of every tile as we argue that it can provide a valuable, rich source of information. Unlike most existing MIL approaches where the relationships developed between neighboring tiles are omitted, in our approach, we propose a neural network architecture that explicitly leverages the dependencies between neighboring tiles of a WSI by enforcing bio-topological constraints to enhance performance effectively. 

\section{Material and methods}\label{sec:methods}

CAMIL operates on the principle that the context and characteristics of a tile's surroundings hold substantial potential for enhancing the accuracy of whole slide classification. To illustrate this concept, we can draw a parallel between our framework and the examination process of a pathologist analyzing a biopsy slide. Similar to how a pathologist inspects sub-regions to comprehensively understand its broader surroundings, CAMIL expands the tile's view to examine the broader neighborhood of each tile thoroughly. This extension allows CAMIL to gather additional information and facilitates a better understanding and assessment of the surrounding microenvironment and tissue context.

In CAMIL, we recalibrate each tile's individual attention score by aggregating the attention scores of its surroundings. For example, tiles with high attention scores surrounded by other high-scoring tiles should be considered important. Conversely, the presence of a tile classified by the model as important in a low-scoring neighborhood could be, in some cases, attributed to noise, and this should be reflected in its final attention score. 

\begin{figure}[t]
\centering
\includegraphics[width=\textwidth]{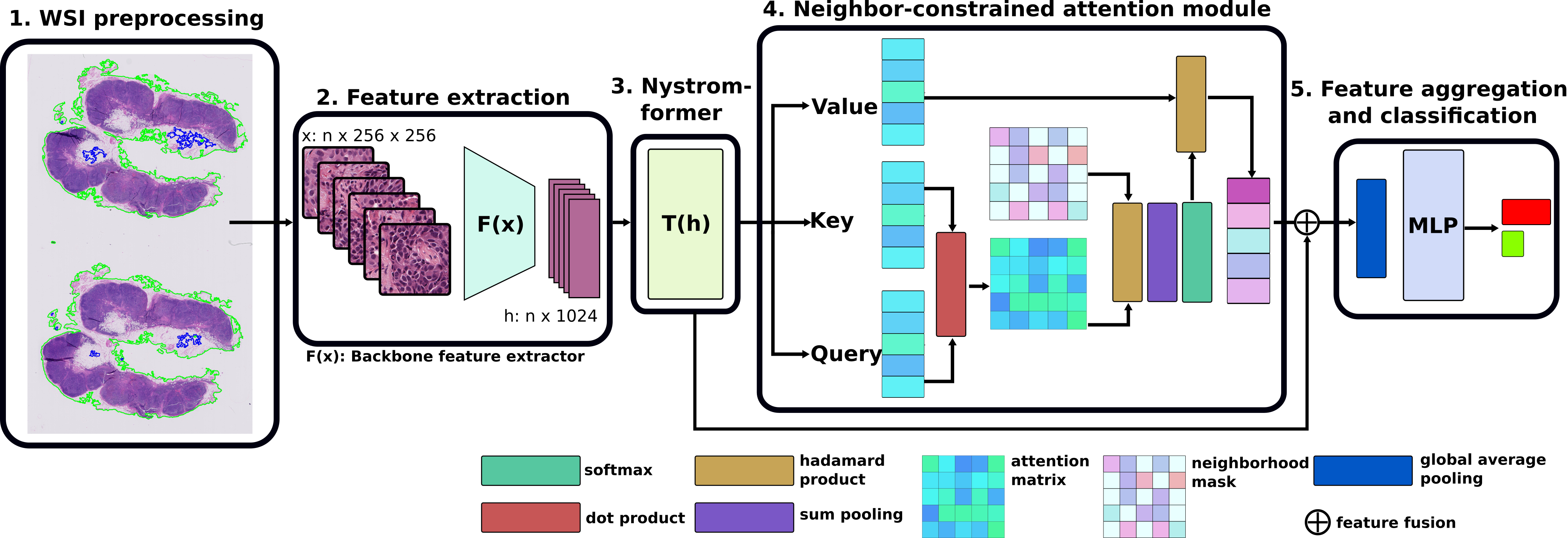}
 \caption {\footnotesize An overview of the CAMIL model architecture. First, WSIs are preprocessed to separate tissue from the background. Then, the WSIs are split into fixed-size tiles of size $256\times256$ and fed through a pre-trained feature extractor to obtain feature representations of size 1024 for each tile. A Nystromformer module then transforms these feature embeddings. These transformed feature embeddings are then used as input to our neighbor-constrained attention module. This module allows attending over each patch and its neighboring patches, generating a neighborhood descriptor of each tile's closest neighbors, and calculating their attention coefficients. The output layer then aggregates the tile-level attention scores produced in the previous layer to emit a final slide classification score.}
 \label{fig:my_model}
 \vspace{-2mm}
\end{figure}

The overview of CAMIL can be seen in Figure \ref{fig:my_model}. It can be decomposed into five elements: 

\begin{enumerate}
    \item A WSI-preprocessing phase automatically segments the tissue region of each WSI and divides it into many smaller tiles (e.g., 256 x 256 pixels).
    \item A tile and feature extraction module, consisting of a stack of convolutional, max pooling, and linear layers transform the original tile input to low dimensional feature representations:  $\boldsymbol{H} = \{\boldsymbol{h}_1, \ldots, \boldsymbol{h}_i, \ldots, \boldsymbol{h}_N\}, \boldsymbol{h}_i \in \mathbb{R}^{n \times d} $, where $d$ is the embedding dimensions of a tile, $n$ the number of tiles within a WSI ($n$ differs among different WSIs), and $N$ the number of WSIs. 
    \item A Nystromformer module \citep{DBLP:conf/aaai/XiongZCTFLS21} transforms the tile embeddings to a concise, descriptive hidden feature representation. It is crucial in aggregating global contexts, capturing the overall information and patterns across multiple tiles.
    \item A neighbor-constrained attention mechanism in CAMIL, coupled with a contrastive learning block, encapsulates the neighborhood prior and focuses on aggregating local concepts. 
    \item  The feature aggregator and classification layer combine the local concepts derived from the previous layer with the features that describe the global contexts obtained from the transformer module. These features are merged to generate a prediction at the slide level.
\end{enumerate}

We elaborate on each step in the following subsections. 

\subsection{WSI preprocessing}
We followed methods in \citet{DBLP:journals/corr/abs-2004-09666} to segment tissue regions and split the whole slide image into individual non-overlapping tiles (1. in Figure \ref{fig:my_model}) (details of the hyperparameters used here are shown in the appendix).

\subsection{Feature extractor}
Effective feature representations significantly impact predictive accuracy, as demonstrated by the success of self-supervised contrastive learning \citep{DBLP:conf/icml/ChenK0H20}. Therefore, to extract rich, meaningful feature representations from individual tiles, we first train a feature extractor following the SimCLR \citep{DBLP:conf/icml/ChenK0H20} approach. SimCLR is one of the most popular self-supervised learning frameworks that enable semantically rich feature representations to be learned by minimizing the distance between different augmented versions of the same image data.

Similar to the training approach followed by \citet{li2021dual}, the data sets utilized in SimCLR are composed of patches derived from WSIs. These patches are densely cropped with no overlap and treated as separate images for the purpose of SimCLR training. During training, two different augmentations are done on the same tile. These two augmentations are chosen from four possible augmentations (color distortion, zoom, rotation, and reflection) using a stochastic data augmentation module. These two augmentations of the same tile are fed through a ResNet-18 \citep{he2015deep} pre-trained on ImageNet \citep{imagenet} with an additional projection head, which is a multi-layer perceptron (MLP) with two hidden layers that map the feature representations to a space where a contrastive loss is applied. The final convolutional block of ResNet-18 and the projection head are then fine-tuned by minimizing the contrastive loss (temperature-scaled cross entropy) between $\boldsymbol{z}_i, \boldsymbol{z}_j$, corresponding to two `correlated' (differentially augmented) views of the same tile. Here, we minimized the normalized temperature-scaled cross entropy (NT-Xent) defined as

\begin{equation}
l_{ij} = -\log \frac{\exp(\mathrm{sim}(\boldsymbol{z}_i,\boldsymbol{z}_j)/\tau)}{\sum_{k=1}^{N} \mathds{1}_{[k \neq  i]} \exp(\mathrm{sim}(\boldsymbol{z}_i,\boldsymbol{z}_k)/\tau)}
\end{equation}

The trained network is then used as the base feature extractor ($F(\boldsymbol{x})$ in Figure 1) to produce the feature representations $\boldsymbol{H} = \{\boldsymbol{h}_1, \ldots, \boldsymbol{h}_i,\ldots,\boldsymbol{h}_N\}, \boldsymbol{h}_i \in \mathbb{R}^{n \times d}$ of each WSI, where  $n$ is the number of tiles and $d$ is the embedding dimension to represent each tile, and $N$ the number of WSIs. This trained feature extractor is frozen when training CAMIL and is only used to extract features and calculate distances between neighboring patches. These distances between neighboring patches are calculated using the sum of squared differences between the features and are used in the following neighbor-constrained attention module.

\subsection{Transformer module to capture global contexts} 
To encode the feature embeddings $\boldsymbol{H}$, our approach focuses on capturing the inter-tile relationships and dependencies, enhancing the global context understanding, and facilitating comprehensive feature aggregation. This is achieved using a transformer layer, represented as $T(\boldsymbol{h})$ in Figure \ref{fig:my_model}, which is particularly effective for managing the complex structure of WSIs. To address the challenge of memory overload due to the long-range dependencies in large WSIs, we adopt the Nystromformer architecture \citep{DBLP:conf/aaai/XiongZCTFLS21}, enabling CAMIL to model intricate feature interactions through an efficient, approximate self-attention mechanism. This produces a ``transformed'' feature set \( \boldsymbol{T}= \{\boldsymbol{t}_1, \ldots , \boldsymbol{t}_i, \ldots ,\boldsymbol{t}_N\} \), with each $\boldsymbol{t}_i \in \mathbb{R}^{n \times d}$, where,

\begin{equation}
\boldsymbol{t}_i = \text{softmax}\Bigg(\frac{\boldsymbol{Q_1}(\boldsymbol{h_i}) \boldsymbol{\Tilde{K}}_1^{T}(\boldsymbol{h_i})} {\sqrt{\text{d}_\text{k}}}\Bigg)
\Bigg( \boldsymbol{A} \Bigg)^+
\text{softmax}\Bigg(\frac{\boldsymbol{\Tilde{Q}}_1(\boldsymbol{h_i}) \boldsymbol{K}_1^{T}(\boldsymbol{h_i})} {\sqrt{\text{d}_\text{k}}}\Bigg)\boldsymbol{V_1}(\boldsymbol{h_i}),
\end{equation}
where $\boldsymbol{\Tilde{Q}}_1(\boldsymbol{h_i})$ and $\boldsymbol{\Tilde{K}}_1(\boldsymbol{h_i})$ are the $m$ selected landmarks (see \citet{DBLP:conf/aaai/XiongZCTFLS21} and Appendix) from the original $n$-dimensional sequence of $\boldsymbol{Q}_1$ and $\boldsymbol{K}_1$, $ \boldsymbol{A}^+=\text{softmax}\Bigg(\frac{\boldsymbol{\Tilde{Q}}_1(\boldsymbol{h_i}) \boldsymbol{\Tilde{K}}_1^{T}(\boldsymbol{h_i})} {\sqrt{\text{d}_\text{k}}}\Bigg)^+ $ is the approximate inverse of $\boldsymbol{A}$. Softmax is applied along the rows of the matrix. $\boldsymbol{K_1}(\boldsymbol{h_i})$, $\boldsymbol{Q_1}(\boldsymbol{h_i})$, and $\boldsymbol{V_1}(\boldsymbol{h_i})$ are the first key, query, and value representations of $\boldsymbol{h_i}$ shown as $T(\boldsymbol{h})$ in Figure \ref{fig:my_model} and defined in \citet{DBLP:conf/nips/VaswaniSPUJGKP17}.

\subsection{Neighbor-constrained attention module to capture local contexts}

The neighbor-constrained attention module in CAMIL is designed to capture specific features and patterns within localized areas of the slide. It focuses on the immediate neighborhood of each tile and aims to capture the local relationships and dependencies within that specific region. By doing so, the module can emphasize the relevance and importance of nearby tiles, effectively incorporating fine-grained details and local nuances into the model.

To model the tile and its surroundings, we construct a weighted adjacency matrix. Consider an undirected graph $G = (V, E)$, where $V$ represents the set of nodes representing image tiles, and $E$ represents the set of edges between nodes indicating adjacency. The graph can be represented by an adjacency matrix \( \boldsymbol{A} \) with elements \( A_{i,j} \), where \( A_{i,j} = s_{ij} \) if there exists an edge \( (v_i, v_j) \in E \) and \( A_{i,j} = 0 \) otherwise. Each image tile must be connected to other tiles and can be surrounded by eight adjacent patches. Each element of the matrix $s_{ij}$ represents the degree of similarity or resemblance between two connected tiles and is calculated as follows:

\begin{equation}
\vspace{-1mm}
\small
s_{ij} = \left\{\begin{array}{ll} 
 \exp {-(\sqrt{(\boldsymbol{h}_i-\boldsymbol{h}_j)^2})} \, , & (v_i, v_j) \in E  \\
 0 \, , & \text{otherwise} \, .
 \end{array} \right.
 \label{eq:adjacency_matrix}
 \end{equation}

This design ensures injecting a bio-topological prior such that the weight of a tile is dependent on adjacent tiles with a similar pattern. 

The transformed tile representations \( \boldsymbol{T}= \{\boldsymbol{t}_1, \ldots , \boldsymbol{t}_i, \ldots ,\boldsymbol{t}_N\} \) are again transformed by the weight matrices \( \boldsymbol{W}_q \in \mathbb{R}^{n \times d_q}\), \( \boldsymbol{W}_k \in \mathbb{R}^{n \times d_k}\) and \( \boldsymbol{W}_v \in \mathbb{R}^{n \times d_v}\) into three distinct representations: the query representation \( \boldsymbol{Q}(\boldsymbol{t}_i)=\boldsymbol{W}_{q}^{\intercal} \boldsymbol{t}_i \), the key representation \( \boldsymbol{K}(\boldsymbol{t}_i)=\boldsymbol{W}_{k}^{\intercal} \boldsymbol{t}_i\) and the value representation \( \boldsymbol{V}(\boldsymbol{t}_i)=\boldsymbol{W}_{v}^{\intercal} \boldsymbol{t}_i \), where \(d_q= d_k= d_v=d\). The dot product of every query with all the key vectors produces an attention matrix whose elements determine the correlation between the different tiles of a WSI (4. in Figure \ref{fig:my_model}).

The similarity mask is element-wise multiplied with the dot product of the query and key embeddings, generating a masked attention matrix whose non-zero elements reflect the contribution of a tile's neighbors to the tile score.

After obtaining the attention coefficients that correspond to the neighbors of every tile, the last step is to aggregate this contextual information to generate a single attention weight. For each tile, we sum the coefficients of their neighbors. The resultant tile score vector is passed through a softmax function to ensure all weights sum to one.

Therefore, the attention coefficient of the \(i\)th tile of a WSI is given by the following equation, where \(\langle.\rangle\) denotes the inner product between two vectors: 
\begin{equation}\label{Score}
\displaystyle w_i =\frac{\exp \bigg( \sum_{j=1}^{N} \langle \boldsymbol{Q}(\boldsymbol{t}_i), \boldsymbol{K}(\boldsymbol{t}_j) \rangle  s_{ij} \bigg)}{\sum_{k=1}^{N} \exp \bigg( \sum_{j=1}^{n} \langle \boldsymbol{Q}(\boldsymbol{t}_k), \boldsymbol{K}(\boldsymbol{t}_j) \rangle  s_{kj} \bigg)} 
\, .
\end{equation}
The feature embeddings \(\boldsymbol{t} \in \mathbb{R}^{n \times d_v}\) are then computed and weighted by their respective attention score  to give a neighbor-constrained feature vector, \(\boldsymbol{l}_i\), for each tile: 

\begin{equation}
\displaystyle \boldsymbol{l}_i=  w_i \boldsymbol{V}(\boldsymbol{t}_i)
\, 
\label{bagLevelRepresentation}
\end{equation}

\subsection{Feature aggregation and slide-level prediction}

The mechanism utilized to fuse the local and global value vectors allows for the adaptive blending of local and global information described in Equation \ref{aff}. The sigmoid function applied to the local values serves as a weighting factor, enabling the model to emphasize local characteristics when they are deemed more relevant while still retaining the contribution from the global contexts.

\begin{equation}
\boldsymbol{m} = \sigma(\boldsymbol{l}) \odot \boldsymbol{l} + (1 - \sigma(\boldsymbol{l})) \odot \boldsymbol{t} ,
\label{aff}
\end{equation} 
where \(\sigma()\) denotes the sigmoid non-linearity.

The collective, WSI-level representation \(\boldsymbol{m} \in \mathbb{R}^{n \times d}\) is adaptively computed as the weighted average of the \(\boldsymbol{z}\) fused vector: 

\begin{equation}
\displaystyle \boldsymbol{z} = \sum_{i=1}^{N} a_i (\boldsymbol{m}_i)
\, ,
\label{output}
\end{equation}

such that: 

\begin{equation}
\displaystyle a_i = \frac {\exp {\boldsymbol{w}^T (\tanh(\boldsymbol{V}\boldsymbol{t}_i^T) \odot \sigma( \boldsymbol{U}\boldsymbol{t}_i^T))}}{\sum_{j=1}^{K} \exp{\boldsymbol{w}^T (\tanh(\boldsymbol{V}\boldsymbol{t}_j^T) \odot \sigma( \boldsymbol{U}\boldsymbol{t}_j^T) )}} 
\,,
\label{a_i}
\end{equation}

where \(\boldsymbol{U}\), \(\boldsymbol{V}\), and \(\boldsymbol{w}\) are learnable parameters, \( \odot \) is an element-wise multiplication, and \(\tanh()\) is the hyperbolic tangent function.

CAMIL achieves a synergistic effect by combining the value vector of the neighbor-constrained attention module with that of the transformer layer. The transformer layer captures global interactions and dependencies across the entire slide, while the neighbor-constrained attention module complements it by capturing local details and context. Together, they enable CAMIL to integrate both local and global perspectives effectively.

Finally, the slide-level prediction is given via the classification layer
\( \boldsymbol{W}_c \in \mathbb{R}^{c \times d} \):

\begin{equation}
\boldsymbol{y}_{\text{slide}} = \boldsymbol{W}_c \cdot \left(\sum_{i} \boldsymbol{z}_i\right)^T
\end{equation}
where \(c\) corresponds to the number of classes and \(\sum\) the sum pooling operation applied on z. The representation obtained from the high-attended patches is used to minimize a cross-entropy loss, and a final classification score is produced.

\section{Experiments and Results}

To demonstrate the performance of CAMIL in capturing informative contextual relationships and improving classification and localization, various experiments were performed on three histopathology datasets: CAMELYON16 \citep{ehteshami_bejnordi_diagnostic_2017}, CAMELYON17 \citep{8447230}, and TCGA-NSCLC. Additional information about the datasets, including details about the training and test sets and our baseline models, can be found in the appendix.

\subsection{Classification performance}

We evaluated the performance of our context-aware pooling operator by comparing its performance other attention-based MIL models, including CLAM-SB, CLAM-MB \citep{DBLP:journals/corr/abs-2004-09666}, TransMIL \citep{DBLP:conf/nips/ShaoBCWZJZ21}, DTFD-MIL \citep{DBLP:conf/cvpr/ZhangMZQYCZ22}, DSMIL \cite{li2021dual} and GTP \citep{DBLP:journals/tmi/ZhengGGBBBK22}. CLAM-SB and CLAM-MB utilize an attention-based pooling operator within the Attention-Based MIL (AB-MIL) framework \citep{DBLP:conf/icml/IlseTW18}. They focus on the features of individual tiles and incorporate a clustering layer to enhance performance further. TransMIL is a transformer-based aggregator operator, DTFD-MIL leverages class activation maps to estimate the probability of an instance being positive under the AB-MIL framework, and DSMIL uses dual instance and bag classifiers to refine predictions. Lastly, GTP combines a graph-based representation of a WSI and a vision transformer.

The results of using CAMIL to classify WSI in the CAMELYON16, TCGA-NSCLC, and CAMELYON17 datasets are presented in Table \ref{tab:results}. The evaluation of the model's performance in all experiments includes the area under the receiver operating characteristic curve (AUC) and the slide-level accuracy (ACC), which is determined by the threshold of 0.5. When implementing our baselines, we fine-tuned the hyperparameters used in the previously published original work to achieve the best performance. The GTP model has the same configurations as the one described in the original paper \citep{DBLP:journals/tmi/ZhengGGBBBK22}. However, for the CAMELYON16 dataset, the batch size ($k$) was set to 2 due to memory limitations.

\begin{table}
    
    \caption{Classification results on CAMELYON16,  TCGA-NSCLC, and CAMELYON17}
    \label{tab:results}
    \begin{center}
    \begin{tabular}{lcccccc}
    \toprule
    & \multicolumn{2}{c}{CAMELYON16} & \multicolumn{2}{c}{TCGA-NSCLC} &  \multicolumn{2}{c}{CAMELYON17} \\
    \cmidrule(lr){2-3} \cmidrule(lr){4-5} \cmidrule(lr){6-7}
    Method & ACC($\uparrow$)  & AUC($\uparrow$) & ACC($\uparrow$) & AUC($\uparrow$) & ACC($\uparrow$) & AUC($\uparrow$)\\
    \midrule
        CLAM-SB             & 0.877\tiny${ 0.029}$  & 0.933\tiny${0.002}$ &0.903\tiny${0.011}$ & 0.972\tiny${0.004}$ &0.802\tiny$0.039$ &0.849\tiny$0.041$\\
        CLAM-MB             & 0.894\tiny${0.010}$  & 0.938\tiny${0.005}$ &0.904\tiny${ 0.010}$  & 0.973\tiny${ 0.004}$ & 0.803\tiny$0.036$& 0.858\tiny$0.046$\\
        TransMIL            & 0.905\tiny${0.005}$ & 0.950\tiny${0.005}$ & 0.905\tiny${0.011}$& 0.974\tiny${0.003}$ & 0.804\tiny$0.021$& 0.873\tiny$0.031$\\
        DTFD-MIL            & 0.889\tiny${0.007}$ &  0.941\tiny${0.005}$ &0.899\tiny${0.010}$  & 0.964\tiny${0.003}$ & 0.797\tiny$0.029$& \bfseries{0.884}\tiny$0.032$\\  
        GTP                 & 0.883\tiny${0.026}$  &  0.921\tiny${0.026}$ &\bfseries{0.916}\tiny${0.015}$  &  0.973\tiny${0.006}$ &0.800\tiny$0.037$ & 0.762\tiny$0.108$\\
        DSMIL               & 0.874\tiny${0.066}$  & 0.949\tiny${0.006}$ & 0.853\tiny$0.031$& 0.954\tiny$0.015$ & 0.815\tiny$0.031$ & 0.863\tiny$0.043$\\
        \midrule
        \bfseries {CAMIL-L} & 0.910\tiny${0.010}$  &  0.953\tiny${0.002}$ &0.914\tiny${0.011}$ &  0.975\tiny${0.004}$&0.828\tiny$0.027$&0.881\tiny$0.031$\\
        \bfseries {CAMIL-G} & 0.891\tiny${0.001}$  &  0.950\tiny${0.009}$ &0.907\tiny${0.012}$  &  0.973\tiny${0.004}$&0.818\tiny$0.039$&0.875\tiny$0.036$\\
        \bfseries {CAMIL} & \bfseries{0.917}\tiny${ 0.006}$ & \bfseries{0.959}\tiny${0.001}$ &\bfseries{0.916}\tiny${0.007}$ & \bfseries{0.975}\tiny${ 0.003}$ &\bfseries{0.843}\tiny$0.024$&0.881\tiny0.039\\
    \bottomrule
    \end{tabular}
    \end{center}
\end{table}

CAMIL outperforms other MIL models in ACC and AUC across CAMELYON16, TCGA-NSCLC, and CAMELYON17 datasets, narrowly trailing DTFD-MIL on CAMELYON17 by 0.003 in AUC. Its effectiveness notably identifies sparse cancerous regions in WSI, where tumor cells are often minimal, such as in the CAMELYON datasets, where tumor cells may account for as little as 5\% of any WSI, which is particularly common in metastatic sites \citep{cheng_computational_2021}. Specifically, CAMIL outperforms the other models on the CAMELYON16 dataset by significant margins, achieving at least 0.9\% better in the AUC and 1.2\% in ACC than the existing models on CAMELYON16 and 3.8\% in ACC on CAMELYON17.

GTP, with its MinCUT pooling layer, which aims to reduce the complexity of self-attention, performs well on TCGA-NSCLC but less so on CAMELYON datasets. Reducing the complexity may make computation more manageable. However, on large and complex datasets such as CAMELYON16 and CAMELYON17, this pooling operation may result in loss of information, particularly in that of fine-grained details. TCGA-NSCLC is a smaller dataset, allowing the MinCUT pooling in GTP to retain sufficient information and remain competitive. 

\subsection{Localization}

\begin{figure}
\centering
\centerline{\includegraphics[width=0.9\textwidth]{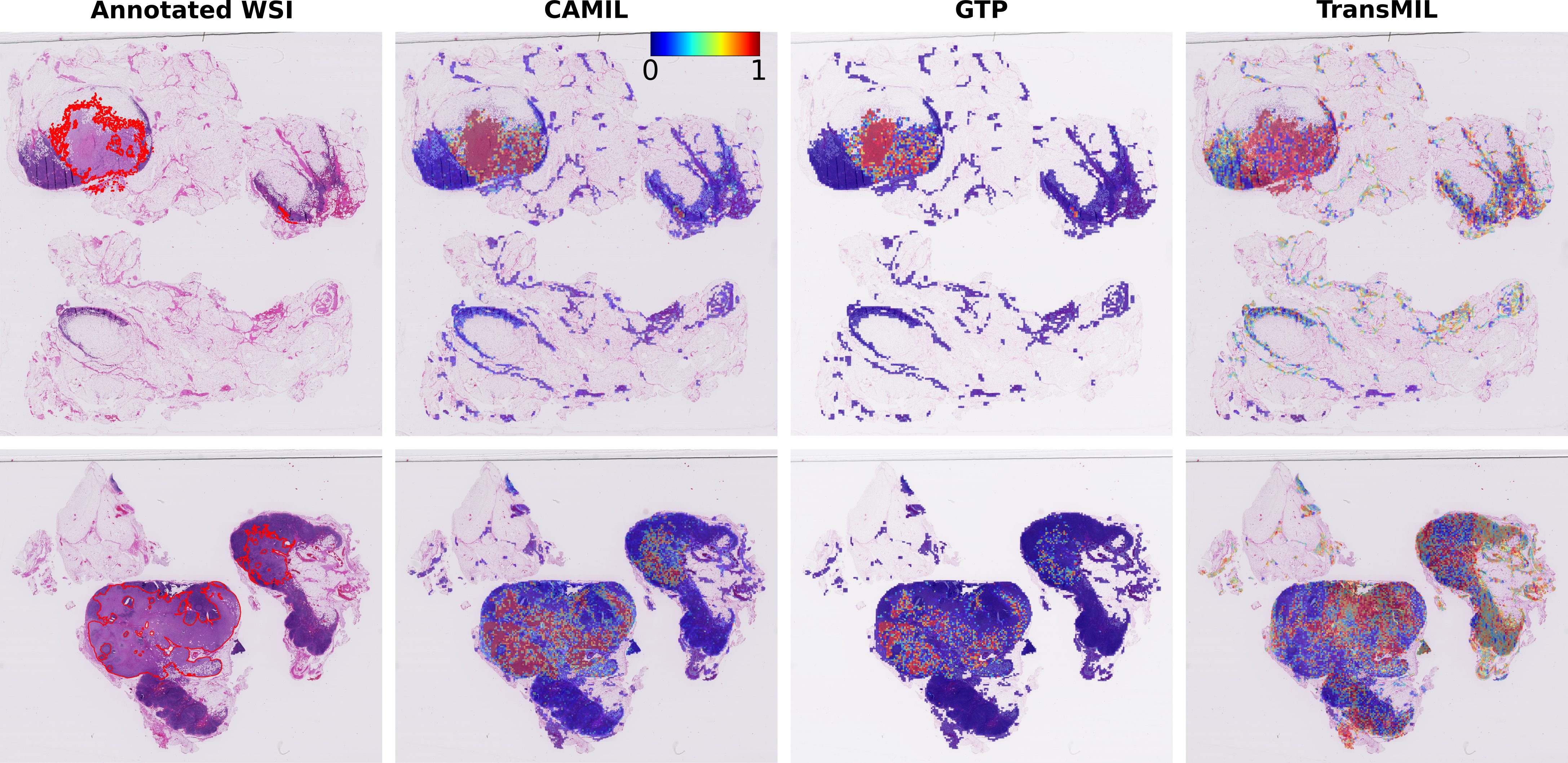}}
 \caption{\footnotesize A visual example of the tumor regions and attention maps produced by different models on the CAMELYON16 cancer dataset. The left column shows the original whole slide image, including the pathologist ground truth annotations with tumor regions delineated by red lines. The other columns show attention maps from left to right for CAMIL, GTP, and TransMIL, pinpointing diagnostically significant locations on the CAMELYON16 cancer dataset. }
 \label{fig:maps_1}
 \end{figure}

 \begin{wraptable}{r}{0.45\textwidth}
    \caption{Localization on CAMELYON16}
    \label{tab:localization}
    \begin{center}
    \begin{tabular}{lcc}
    \toprule
    Method & Dice($\uparrow$)  & Specificity($\uparrow$) \\
    \midrule
        CLAM-SB             &0.459\tiny$0.037$ &0.987\tiny$0.008$\\
        CLAM-MB             & 0.406\tiny$0.007$& 0.573\tiny$0.045$\\
        TransMIL            & 0.103\tiny$0.004$& \bfseries{0.999}\tiny$0.001$\\
        DTFD-MIL            & \bfseries{0.525\tiny$0.033$} & \bfseries{0.999}\tiny$0.001$\\  
        GTP                 & 0.418\tiny$0.068$ & 0.851\tiny$0.116$\\
        DSMIL               & 0.259\tiny$0.083$ & 0.863\tiny$0.043$\\
        \midrule
        \bfseries {CAMIL}  & 0.515\tiny$0.058$&{0.980}\tiny$0.040$\\
    \bottomrule
    \end{tabular}
    \end{center}
\end{wraptable}

To evaluate the localization capability of CAMIL compared to our baselines, we examine both qualitative and quantitative evidence. Similar to the experimental design of \citet{TOURNIAIRE2023102763}, we compute the Dice score to quantify the ability of the different models to identify cancerous evidence in cancerous slides. For normal slides, we compute the tile-level specificity. The reference ground-truth masks are computed at the 5th magnification level using expert tumor delineations. Additionally, a tile is considered cancerous if it contains at least 20\% annotated tumor. To produce the predicted masks, we use the scaled attention scores for CAMIL, both CLAM models, TransMIL and DSMIL, the tile level logits for DTFD-MIL, and the GraphCAM for GTP. We apply a threshold of 0.5 to the model's output probabilities to generate the masks from the tile-level predictions. The results for the Dice score and Specificity are shown in Table \ref{tab:localization}

CAMIL performs well, albeit slightly behind the DTFD-MIL model. We believe the decreased localization performance might be attributed to integrating the Nystromformer module in our model design, akin to its role in TransMIL. TransMIL, as indicated by the attention maps in our qualitative assessment and its slide-level performance, demonstrates the ability to grasp the general patterns within a WSI and distinguish between normal and cancerous slides. However, despite this, confirmed by its low Dice score, it falls short in effectively pinpointing specific cancerous evidence within slides. Integrating the Nystromformer into our model design might introduce a trade-off between slide-level accuracy and localization performance, resulting in improved slide-level accuracy with an expense of slightly decreased localization performance.

Figure~\ref{fig:maps_1} provides a qualitative comparison of the attention maps generated by CAMIL, GTP, and TransMIL on the CAMELYON16 cancer dataset. We visually compare these methods as they leverage spatial information to enhance prediction. These attention maps underscore CAMIL's high localization performance, as it can discern the boundaries separating normal tissue from tumor tissue. Although effectively pinpointing the regions of interest, GTP attention maps appear fragmented and less dense in cancer-associated regions. This fragmentation could be attributed to the MinCUT-pooling operation, which may reduce the representation's granularity and affect the heatmap's coherence. 

TransMIL appears to sufficiently capture long-term dependencies within the WSIs, as evidenced by the attention maps of Figure~\ref{fig:maps_1}. Specifically, TransMIL can identify the presence of cancer and precisely pinpoint the cancer-associated regions. However, these maps also reveal TransMIL's inability to capture intricate details and local nuances. The attention scores are not only confined to the cancer regions but expand beyond those to the surrounding normal tissue, impeding the precise localization of tumor boundaries, indicating the model's limitations in representing close proximity relationships within the WSIs.

\section{Ablation studies}\label{sec:discussion}

Additionally, we performed ablation studies to evaluate the effectiveness of the Nystromformer module and that of the neighboring-constrained attention module in our model. Specifically, we examined the effect of the Nystromformer block by retaining it while excluding the neighbor-constrained attention module denoted as CAMIL-G. Table \ref{tab:results} demonstrates that using only the Nystromformer block leads to satisfactory performance comparable to that of the TransMIL model, which also incorporates the Nystromformer. In a distinct ablation study, we omitted the Nystromformer block and retained the neighbor-constrained attention module, referred to as CAMIL-L. CAMIL-L exhibits a marginal improvement over the CAMIL-G model, thereby underlining its crucial role in augmenting the model's performance. Optimal results were achieved through the amalgamation of both models, harnessing both the potential of the Nystromformer block to model long-term dependencies and the prowess of the neighbor-attention module to comprehend local visual concepts.

\subsection{Visualising global and local concepts}

\begin{figure}
    \centering
    \begin{minipage}[b]{0.545\textwidth}
        \centering
        \includegraphics[width=\textwidth]{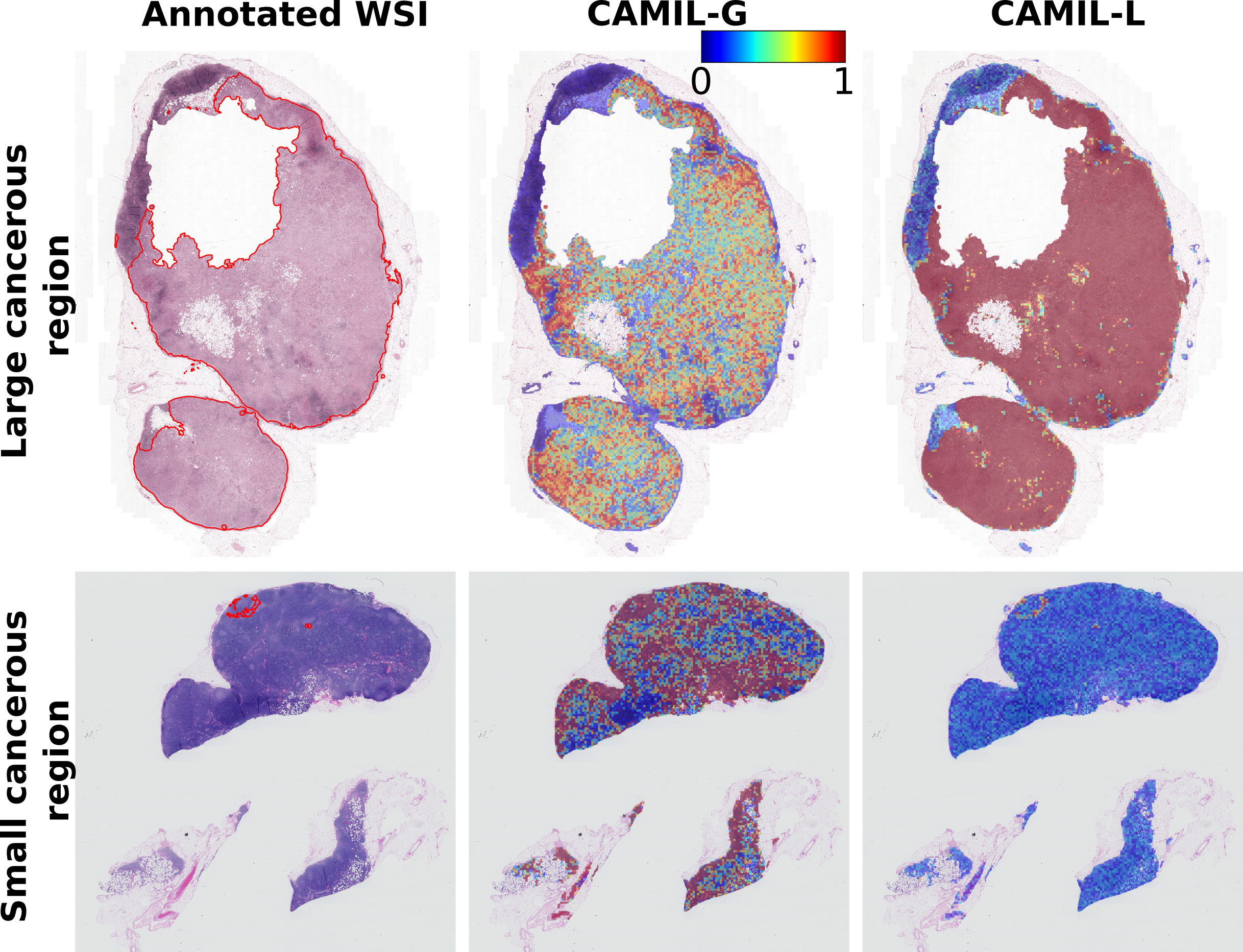}
        \caption{\footnotesize A visual example of the tumor regions and attention maps produced by different models on the CAMELYON16 cancer dataset. From left to right are the ground truth annotations with tumor regions delineated by red lines, the heatmaps for CAMIL-G, and the heatmaps for CAMIL-L.}
        \label{fig:global_maps}
    \end{minipage}
    \hfill
    \begin{minipage}[b]{0.435\textwidth}
        \centering
        \includegraphics[width=\textwidth]{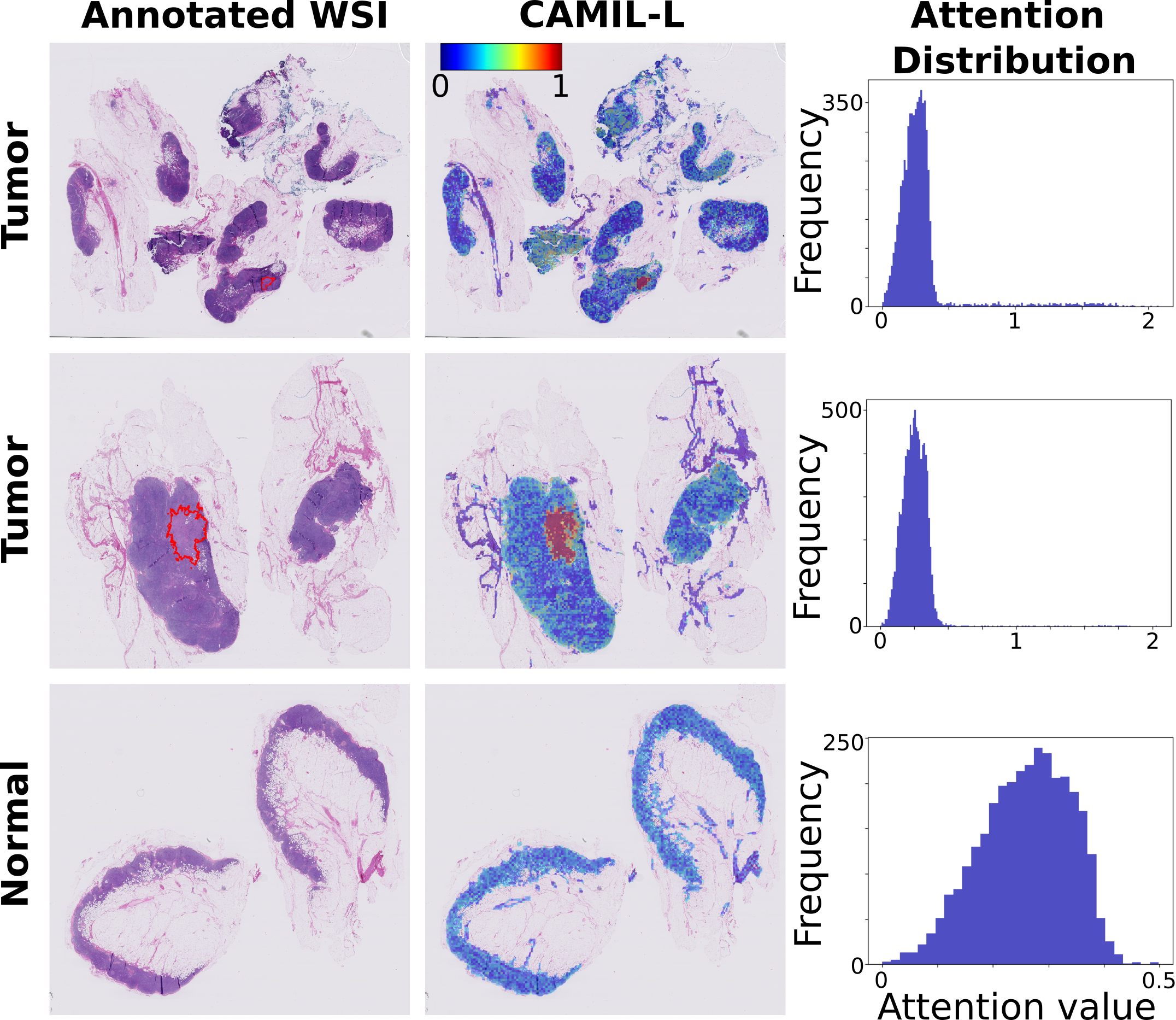}
        \caption{\footnotesize A visual example CAMIL-L fine-grained localization abilities. From left to right are the ground truth annotations with tumor regions delineated by red lines, generated heatmaps for CAMIL-LCAMIL-L, and the attention distribution of the neighbor-constrained attention module's weights (before applying the softmax function).}
        \label{fig:local_maps}
    \end{minipage}
\end{figure}

We also visualized the attention maps generated by the two versions of our model. Notably, in Figure \ref{fig:global_maps}, CAMIL-G demonstrates commendable performance in understanding global concepts and overall patterns, as it effectively detects tumor regions. However, it struggles with highly localized tumors, encountering difficulties capturing intricate, short-term dependencies within the image.

On the other hand, the attention maps produced by CAMIL-L excel in capturing the fine details within a WSI. This proficiency can be attributed to the context-aware module, which utilizes a similarity mask. This mask aggregates attention weights of similar neighboring feature representations, resulting in robust activations. In contrast, less favorable weights, particularly those associated with negative regions, do not contribute as strongly. This observation is substantiated by the histograms of the unnormalized attention coefficients, which are not constrained within the range of 0 to 1 (before applying the softmax function) provided in Figure \ref{fig:local_maps}. These histograms underscore a notable pattern: while a significant portion of the attention coefficients falls below the 0.5 threshold, cancerous cases display outliers within the range of 0.5 to 2. These outliers represent stronger activations, which distinctly characterize regions affected by cancer. Conversely, all other activations consistently remain below the 0.5 threshold, signifying a reduced emphasis on non-cancerous areas.

\section{Conclusion}\label{sec:conclusion}

We have introduced CAMIL, a novel MIL vision transformer-based method that considers the tumor microenvironment context while determining tile-level labels in WSIs, mirroring the approach of a skilled pathologist. This is achieved by employing a unique neighbor-constrained attention mechanism, which assesses the dependencies between tiles within a WSI and incorporates contextual constraints as prior knowledge into the MIL model. We have demonstrated that using the transformer and the neighborhood-attention mechanism together is imperative in successful performance across datasets through our ablation studies. Importantly, CAMIL achieves state-of-the-art across multiple datasets regarding tile-level ACC, AUC, and F1 scores and patch-level localization and interpretability. 


\bibliography{iclr2024_conference}
\bibliographystyle{iclr2024_conference}

\appendix
\section{Appendix}

\subsection{Reproducibility statement}
All weakly-supervised deep learning models are trained using NVIDIA Tesla P100 GPUs. One GPU is used for training in each experiment. We intend to make the source code of our algorithm publicly available soon.

\subsection{Performance metrics}
Here, we provide ACC, F1, and AUC for both datasets. F1 scores were omitted from the main text to save space.
\begin{table}[H]
    \centering
    \caption{Classification results on CAMELYON16}
    \label{tab:appendix_tab_camel}
    \begin{tabular}{rccc}
    METHOD & ACC($\uparrow$) & $F_1$($\uparrow$) & AUC($\uparrow$)\\
\hline & \\[-1.5ex]

\bfseries {CLAM-SB} & 0.877\tiny${ 0.029}$ & 0.840\tiny${0.023}$ & 0.933\tiny${0.002}$ \\
\bfseries {CLAM-MB} & 0.894\tiny${0.010}$ & 0.845 \tiny${0.019}$ & 0.938\tiny${0.005}$\\
\bfseries {TransMIL} & 0.905\tiny${0.005}$  & \bfseries{0.888}\tiny${0.018}$ & 0.950\tiny${0.005}$ \\
\bfseries {DTFD-MIL} & 0.889\tiny${0.007}$ & 0.850\tiny${0.004}$ &  0.941\tiny${0.005}$\\
\bfseries {GTP} & 0.883\tiny${0.026}$ & 0.863\tiny${0.026}$ &  0.921\tiny${0.026}$\\
\bfseries{DSMIL} & 0.874\tiny$0.066$& 0.848\tiny$0.056$ & 0.949\tiny$0.006$\\
\hline & \\[-1.5ex]
\bfseries {CAMIL-L} & 0.910\tiny${0.010}$ & 0.872\tiny${0.016}$ &  0.953\tiny${0.002}$\\
\bfseries {CAMIL-G} & 0.891\tiny${0.001}$ & 0.866\tiny${0.012}$ &  0.950\tiny${0.009}$\\
\bfseries {CAMIL}& \bfseries{0.917\tiny${ 0.006}$} & 0.881\tiny${0.010}$ & \bfseries{0.959}\tiny${0.001}$\\
\hline
    \end{tabular}
\end{table}

\begin{table}[H]
    \centering
    \caption{Classification results on TCGA-NSCLC}
    \label{tab:appendix_table_tcga}
    \begin{tabular}{rccc}
         \hspace*{-10cm} 
 METHOD & ACC($\uparrow$) & $F_1$($\uparrow$) & AUC($\uparrow$)\\
\hline & \\[-1.5ex]
\bfseries {CLAM-SB} &0.903\tiny${0.011}$ & 0.867 \tiny${0.012}$ & 0.972\tiny${0.004}$\\
\bfseries {CLAM-MB}  &0.904\tiny${ 0.010}$ & 0.872\tiny${0.057}$ & 0.973\tiny${ 0.004}$ \\
\bfseries {TransMIL} &0.905\tiny${0.011}$ &0.911\tiny${  0.005}$ & 0.974\tiny${0.003}$ \\
\bfseries {DTFD-MIL}   &0.899\tiny${0.010}$ & 0.899\tiny${0.014}$ & 0.964\tiny${0.003}$ \\

\bfseries {GTP} &\bfseries{0.916}\tiny${0.015}$ &0.917\tiny${0.016}$ &  0.973\tiny${0.006}$\\
\bfseries{DSMIL} & 0.853\tiny$0.031$& 0.864\tiny$0.024$& 0.954\tiny$0.015$ \\ 
\hline & \\[-1.5ex]
 \bfseries {CAMIL-L}  &0.914\tiny${0.011}$ &\bfseries{0.921} \tiny${0.009}$ &  0.975\tiny${0.004}$\\
\bfseries {CAMIL-G} &0.907\tiny${0.012}$ & 0.918\tiny${0.007}$ &  0.973\tiny${0.004}$\\
\bfseries {CAMIL} & \bfseries{0.916}\tiny${0.007}$ & 0.918\tiny${0.005}$ & \bfseries{0.975}\tiny${ 0.003}$ \\
\hline
    \end{tabular}
\end{table}

\subsection{WSI pre-processing}\label{data_pre_processing}

Using the publicly available WSI-prepossessing toolbox developed by \citep{DBLP:journals/corr/abs-2004-09666}, for both datasets, we first automatically segmented the tissue region from each slide and exhaustively divided it into 256$\times$256 non-overlapping patches using $\times 20$ magnification (Figure 1). Otsu's method was used to perform automatic WSI thresholding.

To avoid the computational overhead and capitalize on the rich feature representations already learned during its previous training on CAMELYON16, CAMELYON17, and TCGA-NSCLC datasets, we opted to use the pre-trained ResNet-18 feature extractor provided by \citep{li2021dual}. This model was extensively trained on a large set of tiles from the CAMELYON16 dataset, densely cropped without overlap, making it a powerful feature extractor. To make a fair comparison, we used the same contrastive learning-based model as the feature extractor for all our baselines.

\subsection{Datasets}\label{sec:results}

CAMELYON16 is a significant publicly available Whole Slide Image (WSI) dataset for lymph node classification and metastasis detection. It includes 270 training and 129 test slides from two medical centers, all meticulously annotated by pathologists. Some slides have partial annotations, making it a challenging benchmark due to varying metastasis sizes.

The CAMELYON17 dataset consists of 1000 WSIs of a similar type to the CAMELYON16 dataset. However, only half (500 WSI) of these images are labeled and accessible publicly. These images are expertly annotated by pathologic lymph node classification into pN-stage:
\begin{itemize}
    \item pN0: No micro-metastases or macro-metastases, or isolated tumor cells (ITCs) found.
    \item pN0(i+): Only ITCs found.
    \item pN1mi: Micro-metastases found, but no macro-metastases found.
    \item pN1: Metastases found in 1–3 lymph nodes, of which at least one is a macro-metastasis.
    \item pN2: Metastases found in 4–9 lymph nodes, of which at least one is a macro-metastasis.
\end{itemize}

Adopting the approach discussed in \citet{TOURNIAIRE2023102763}, we created a binary classification problem by treating pN0 as normal and unifying all classes that were not pN0 into a single class, cancerous.

The TCGA-NSCLC dataset comprises two non-small cell lung cancer subtypes, LUAD and LUSC, with 541 slides. Unlike CAMELYON16, it lacks annotations.

\subsection{Data splits}

In the case of CAMELYON16, the WSIs are partitioned into a training and test set. The 270 WSIs of the training set are split five times into a training (80\%) and a validation (20\%) set in a 5-fold cross-validation fashion, and the average performance of the model on the competition test set is reported. The official test set comprising 129 WSIs is used for evaluation. For CAMELYON17, we used a 4-fold validation 65\%, 15\%, and 25\% train, validation, and test splits. Regarding the TCGA-NSCLC dataset, a 5-fold cross-validation across the available images is performed. For each fold, the training set is split into 80\% for training purposes and 20\% for validation.

\subsection{Transformer module to capture global contexts} 

Working with large WSIs can lead to memory overload, as the self-attention mechanism used in the transformer layer requires computing pairwise interactions between all of the tiles in each WSI. To circumvent the memory overload associated with the long-range dependencies of large WSIs, we adopt the Nystromformer architecture \citep{DBLP:conf/aaai/XiongZCTFLS21} to model feature interactions that otherwise would be intractable. 

The Nystromformer approach is based on the Nystrom method, which is a technique for approximating a kernel matrix by selecting a small subset of its rows and columns. In the context of the transformer layer, this means selecting a subset of "landmark" tiles from each WSI to represent the full set of tiles. The landmark tiles are chosen randomly, and their embeddings are used to compute a low-rank approximation of the self-attention matrix. This approximation is then used instead of the full self-attention matrix to compute the final output of the transformer layer.

The Nystromformer architecture is highly scalable regarding sequence length, making it well-suited for processing large WSIs  \citep{DBLP:conf/aaai/XiongZCTFLS21}. Additionally, by reducing the time complexity of the self-attention mechanism from $\mathcal{O}(n^2)$ to $\mathcal{O}(n)$, the Nystromformer approach can significantly reduce the computational cost of processing each WSI.

\subsection{Localisation} 

 \begin{figure}[hbt!]
\centering
\centerline{\includegraphics[width=\textwidth]{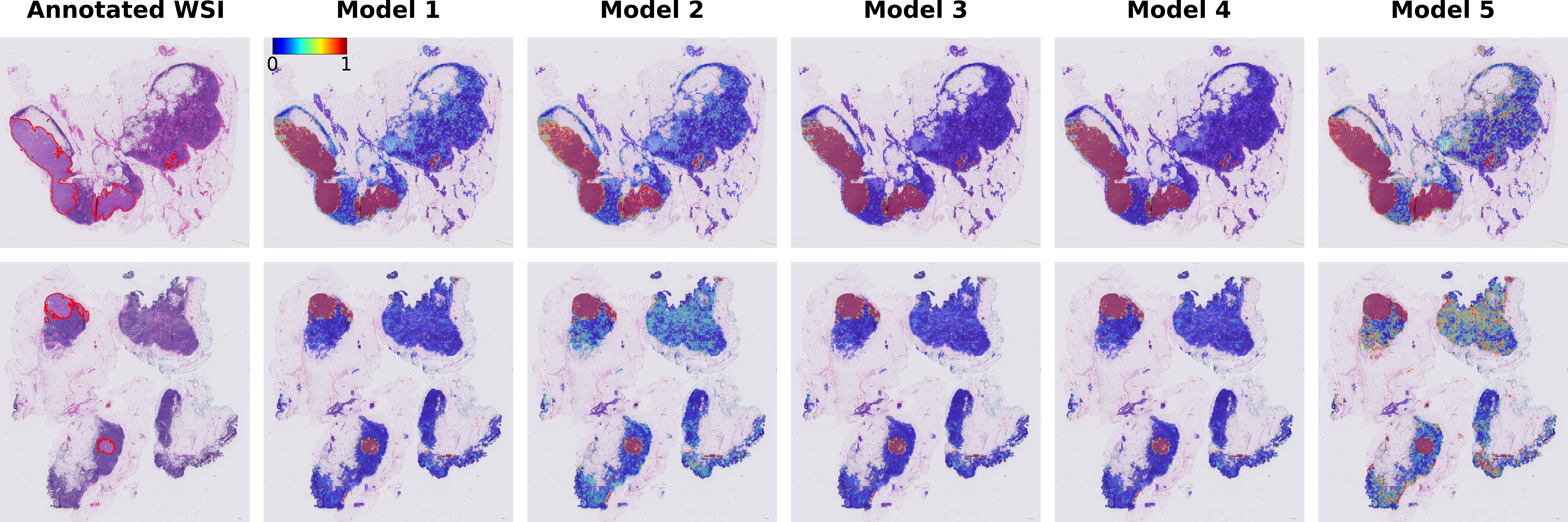}}
 \caption{\footnotesize Displayed are attention maps produced on Whole Slide Images (WSIs) throughout the 5-fold cross-validation runs. The leftmost column shows the original WSIs, including ground truth annotations of tumor regions delineated by red lines, while the succeeding columns depict the predicted probabilities from the cross-validated model runs. The colormap signifies the likelihood with which a specific region within a WSI correlates to the target output label.}
 \label{fig:maps_2}
 \end{figure}
 
In addition to the quantitative outcomes, we assessed our approach qualitatively by visually presenting attention maps generated by different validation runs of our model overlaid on expert-annotated tumor regions (Figure \ref{fig:maps_2}). These attention maps pinpoint diagnostically significant locations in the image that are crucial for accurate tumor identification. 

We notice a substantial agreement between the regions of interest identified by experts and those generated by our attention maps. These steps are harmonized with an attention map, forming a transformer relevancy map. Notably, our method consistently highlights the same regions within Whole Slide Images (WSIs) across different cross-validation folds, underscoring the reliability and robustness of our model.

\subsection{SimCLR ablation study}
SimCLR feature extraction backbone is an integral part of our model, and it plays a pivotal role in generating concise and descriptive feature representations. Leveraging the pre-trained SimCLR weights lays the foundation for generating meaningful similarity scores and, therefore, is crucial for the optimal performance of the neighbor-constrained attention mask similar to many similarity-based histopathology approaches. Incorporating ImageNet weights directly into our model notably decreases its performance (as it does with most models \cite{DBLP:journals/mia/TourniaireIHAD23, li2021dual}), emphasizing the necessity of SimCLR within our model architecture. Table \ref{tab:resnet_albation} shows these results when using a ResNet-18 pre-trained on ImageNet on CAMELYON16 and TCGA-NSCLC.

\begin{table}[]
    \centering
    \caption{CAMIL using ResNet18 feature extractor pre-trained on ImageNet without SimCLR fine-tuning.}
    \label{tab:resnet_albation}
    \begin{tabular}{c|c|c}

    Dataset & ACC & AUC \\
    \hline
    CAMELYON16 & 0.723\tiny$0.095$ & 0.743\tiny$0.077$\\
    TCGA-NSCLC &0.692\tiny$0.082$& 0.798\tiny$0.059$\\ 
       \hline
    \end{tabular}
    
\end{table}

\end{document}

%% file: math_commands.tex
\usepackage{amsmath,amsfonts,bm}

\def\eqref#1{equation~\ref{#1}}

\def\1{\bm{1}}

\DeclareMathAlphabet{\mathsfit}{\encodingdefault}{\sfdefault}{m}{sl}
\SetMathAlphabet{\mathsfit}{bold}{\encodingdefault}{\sfdefault}{bx}{n}













%% file: iclr2024_conference.bbl
\begin{thebibliography}{44}
\providecommand{\natexlab}[1]{#1}
\providecommand{\url}[1]{\texttt{#1}}
\expandafter\ifx\csname urlstyle\endcsname\relax
  \providecommand{\doi}[1]{doi: #1}\else
  \providecommand{\doi}{doi: \begingroup \urlstyle{rm}\Url}\fi

\bibitem[Adnan et~al.(2020)Adnan, Kalra, and Tizhoosh]{DBLP:conf/cvpr/AdnanKT20}
Mohammed Adnan, Shivam Kalra, and Hamid~R. Tizhoosh.
\newblock Representation learning of histopathology images using graph neural networks.
\newblock In \emph{2020 {IEEE/CVF} Conference on Computer Vision and Pattern Recognition, {CVPR} Workshops 2020, Seattle, WA, USA, June 14-19, 2020}, pp.\  4254--4261. Computer Vision Foundation / {IEEE}, 2020.
\newblock \doi{10.1109/CVPRW50498.2020.00502}.

\bibitem[BenTaieb \& Hamarneh(2018)BenTaieb and Hamarneh]{DBLP:conf/miccai/BenTaiebH18}
A{\"{\i}}cha BenTaieb and Ghassan Hamarneh.
\newblock Predicting cancer with a recurrent visual attention model for histopathology images.
\newblock In Alejandro~F. Frangi, Julia~A. Schnabel, Christos Davatzikos, Carlos Alberola{-}L{\'{o}}pez, and Gabor Fichtinger (eds.), \emph{Medical Image Computing and Computer Assisted Intervention - {MICCAI} 2018 - 21st International Conference, Granada, Spain, September 16-20, 2018, Proceedings, Part {II}}, volume 11071 of \emph{Lecture Notes in Computer Science}, pp.\  129--137. Springer, 2018.
\newblock \doi{10.1007/978-3-030-00934-2\_15}.

\bibitem[Bándi et~al.(2019)Bándi, Geessink, Manson, Van~Dijk, Balkenhol, Hermsen, Ehteshami~Bejnordi, Lee, Paeng, Zhong, Li, Zanjani, Zinger, Fukuta, Komura, Ovtcharov, Cheng, Zeng, Thagaard, Dahl, Lin, Chen, Jacobsson, Hedlund, Çetin, Halıcı, Jackson, Chen, Both, Franke, Küsters-Vandevelde, Vreuls, Bult, van Ginneken, van~der Laak, and Litjens]{8447230}
Péter Bándi, Oscar Geessink, Quirine Manson, Marcory Van~Dijk, Maschenka Balkenhol, Meyke Hermsen, Babak Ehteshami~Bejnordi, Byungjae Lee, Kyunghyun Paeng, Aoxiao Zhong, Quanzheng Li, Farhad~Ghazvinian Zanjani, Svitlana Zinger, Keisuke Fukuta, Daisuke Komura, Vlado Ovtcharov, Shenghua Cheng, Shaoqun Zeng, Jeppe Thagaard, Anders~B. Dahl, Huangjing Lin, Hao Chen, Ludwig Jacobsson, Martin Hedlund, Melih Çetin, Eren Halıcı, Hunter Jackson, Richard Chen, Fabian Both, Jörg Franke, Heidi Küsters-Vandevelde, Willem Vreuls, Peter Bult, Bram van Ginneken, Jeroen van~der Laak, and Geert Litjens.
\newblock From detection of individual metastases to classification of lymph node status at the patient level: The camelyon17 challenge.
\newblock \emph{IEEE Transactions on Medical Imaging}, 38\penalty0 (2):\penalty0 550--560, 2019.
\newblock \doi{10.1109/TMI.2018.2867350}.

\bibitem[Campanella et~al.(2019)Campanella, Hanna, Geneslaw, Miraflor, Silva, Busam, Brogi, Reuter, Klimstra, and Fuchs]{Campanella2019ClinicalgradeCP}
Gabriele Campanella, Matthew~G. Hanna, Luke Geneslaw, Allen~P. Miraflor, Vitor Werneck~Krauss Silva, Klaus~J. Busam, Edi Brogi, Victor~E. Reuter, David~S. Klimstra, and Thomas~J. Fuchs.
\newblock Clinical-grade computational pathology using weakly supervised deep learning on whole slide images.
\newblock \emph{Nature Medicine}, pp.\  1--9, 2019.

\bibitem[Chen et~al.(2020)Chen, Kornblith, Norouzi, and Hinton]{DBLP:conf/icml/ChenK0H20}
Ting Chen, Simon Kornblith, Mohammad Norouzi, and Geoffrey~E. Hinton.
\newblock A simple framework for contrastive learning of visual representations.
\newblock In \emph{Proceedings of the 37th International Conference on Machine Learning, {ICML} 2020, 13-18 July 2020, Virtual Event}, volume 119 of \emph{Proceedings of Machine Learning Research}, pp.\  1597--1607. PMLR, 2020.

\bibitem[Cheng et~al.(2021)Cheng, Liu, Huang, Hong, Wang, Zhan, Han, Ni, Huang, and Zhang]{cheng_computational_2021}
Jun Cheng, Yuting Liu, Wei Huang, Wenhui Hong, Lingling Wang, Xiaohui Zhan, Zhi Han, Dong Ni, Kun Huang, and Jie Zhang.
\newblock Computational {Image} {Analysis} {Identifies} {Histopathological} {Image} {Features} {Associated} {With} {Somatic} {Mutations} and {Patient} {Survival} in {Gastric} {Adenocarcinoma}.
\newblock \emph{Frontiers in Oncology}, 11, 2021.
\newblock ISSN 2234-943X.

\bibitem[Courtiol et~al.(2018)Courtiol, Tramel, Sanselme, and Wainrib]{DBLP:journals/corr/abs-1802-02212}
Pierre Courtiol, Eric~W. Tramel, Marc Sanselme, and Gilles Wainrib.
\newblock Classification and disease localization in histopathology using only global labels: {A} weakly-supervised approach.
\newblock \emph{CoRR}, abs/1802.02212, 2018.

\bibitem[Das et~al.(2018)Das, Conjeti, Roy, Chatterjee, and Sheet]{DBLP:conf/isbi/DasCRCS18}
Kausik Das, Sailesh Conjeti, Abhijit~Guha Roy, Jyotirmoy Chatterjee, and Debdoot Sheet.
\newblock Multiple instance learning of deep convolutional neural networks for breast histopathology whole slide classification.
\newblock In \emph{15th {IEEE} International Symposium on Biomedical Imaging, {ISBI} 2018, Washington, DC, USA, April 4-7, 2018}, pp.\  578--581. {IEEE}, 2018.
\newblock \doi{10.1109/ISBI.2018.8363642}.

\bibitem[Deng et~al.(2009)Deng, Dong, Socher, Li, Li, and Fei-Fei]{imagenet}
Jia Deng, Wei Dong, Richard Socher, Li-Jia Li, Kai Li, and Li~Fei-Fei.
\newblock Imagenet: A large-scale hierarchical image database.
\newblock In \emph{2009 IEEE Conference on Computer Vision and Pattern Recognition}, pp.\  248--255, 2009.
\newblock \doi{10.1109/CVPR.2009.5206848}.

\bibitem[Ehteshami~Bejnordi et~al.(2017)Ehteshami~Bejnordi, Veta, Johannes~van Diest, van Ginneken, Karssemeijer, Litjens, van~der Laak, and {and the CAMELYON16 Consortium}]{ehteshami_bejnordi_diagnostic_2017}
Babak Ehteshami~Bejnordi, Mitko Veta, Paul Johannes~van Diest, Bram van Ginneken, Nico Karssemeijer, Geert Litjens, Jeroen A. W.~M. van~der Laak, and {and the CAMELYON16 Consortium}.
\newblock Diagnostic {Assessment} of {Deep} {Learning} {Algorithms} for {Detection} of {Lymph} {Node} {Metastases} in {Women} {With} {Breast} {Cancer}.
\newblock \emph{JAMA}, 318\penalty0 (22):\penalty0 2199--2210, December 2017.
\newblock ISSN 0098-7484.
\newblock \doi{10.1001/jama.2017.14585}.
\newblock URL \url{https://doi.org/10.1001/jama.2017.14585}.

\bibitem[Gadermayr et~al.(2023)Gadermayr, Koller, Tschuchnig, Stangassinger, Kreutzer, Couillard-Despres, Oostingh, and Hittmair]{10.1007/978-3-031-43987-2_46}
Michael Gadermayr, Lukas Koller, Maximilian Tschuchnig, Lea~Maria Stangassinger, Christina Kreutzer, Sebastien Couillard-Despres, Gertie~Janneke Oostingh, and Anton Hittmair.
\newblock Mixup-mil: Novel data augmentation for multiple instance learning and a study on thyroid cancer diagnosis.
\newblock In Hayit Greenspan, Anant Madabhushi, Parvin Mousavi, Septimiu Salcudean, James Duncan, Tanveer Syeda-Mahmood, and Russell Taylor (eds.), \emph{Medical Image Computing and Computer Assisted Intervention -- MICCAI 2023}, pp.\  477--486, Cham, 2023. Springer Nature Switzerland.
\newblock ISBN 978-3-031-43987-2.

\bibitem[He et~al.(2015)He, Zhang, Ren, and Sun]{he2015deep}
Kaiming He, Xiangyu Zhang, Shaoqing Ren, and Jian Sun.
\newblock Deep residual learning for image recognition, 2015.

\bibitem[Hou et~al.(2016)Hou, Samaras, Kur{\c{c}}, Gao, Davis, and Saltz]{DBLP:conf/cvpr/HouSKGDS16}
Le~Hou, Dimitris Samaras, Tahsin~M. Kur{\c{c}}, Yi~Gao, James~E. Davis, and Joel~H. Saltz.
\newblock Patch-based convolutional neural network for whole slide tissue image classification.
\newblock In \emph{2016 {IEEE} Conference on Computer Vision and Pattern Recognition, {CVPR} 2016, Las Vegas, NV, USA, June 27-30, 2016}, pp.\  2424--2433. {IEEE} Computer Society, 2016.
\newblock \doi{10.1109/CVPR.2016.266}.

\bibitem[Ilse et~al.(2018)Ilse, Tomczak, and Welling]{DBLP:conf/icml/IlseTW18}
Maximilian Ilse, Jakub~M. Tomczak, and Max Welling.
\newblock Attention-based deep multiple instance learning.
\newblock In Jennifer~G. Dy and Andreas Krause (eds.), \emph{Proceedings of the 35th International Conference on Machine Learning, {ICML} 2018, Stockholmsm{\"{a}}ssan, Stockholm, Sweden, July 10-15, 2018}, volume~80 of \emph{Proceedings of Machine Learning Research}, pp.\  2132--2141, 2018.

\bibitem[Landini et~al.(2020)Landini, Martinelli, and Piccinini]{10.1093/bioinformatics/btaa847}
Gabriel Landini, Giovanni Martinelli, and Filippo Piccinini.
\newblock {Colour deconvolution: stain unmixing in histological imaging}.
\newblock \emph{Bioinformatics}, 09 2020.
\newblock ISSN 1367-4803.
\newblock \doi{10.1093/bioinformatics/btaa847}.
\newblock btaa847.

\bibitem[Li et~al.(2021{\natexlab{a}})Li, Li, and Eliceiri]{li2021dual}
Bin Li, Yin Li, and Kevin~W Eliceiri.
\newblock Dual-stream multiple instance learning network for whole slide image classification with self-supervised contrastive learning.
\newblock In \emph{Proceedings of the IEEE/CVF Conference on Computer Vision and Pattern Recognition}, pp.\  14318--14328, 2021{\natexlab{a}}.

\bibitem[Li et~al.(2021{\natexlab{b}})Li, Li, Sisk, Ye, Wallace, Speier, and Arnold]{DBLP:journals/cbm/LiLSYWSA21}
Jiayun Li, Wenyuan Li, Anthony~E. Sisk, Huihui Ye, W.~Dean Wallace, William Speier, and Corey~W. Arnold.
\newblock A multi-resolution model for histopathology image classification and localization with multiple instance learning.
\newblock \emph{Comput. Biol. Medicine}, 131:\penalty0 104253, 2021{\natexlab{b}}.
\newblock \doi{10.1016/j.compbiomed.2021.104253}.

\bibitem[Liu et~al.(2023)Liu, Ji, Zhang, and Ye]{liu2023pseudobag}
Pei Liu, Luping Ji, Xinyu Zhang, and Feng Ye.
\newblock Pseudo-bag mixup augmentation for multiple instance learning-based whole slide image classification, 2023.

\bibitem[Lu et~al.(2021)Lu, Williamson, Chen, Chen, Barbieri, and Mahmood]{DBLP:journals/corr/abs-2004-09666}
Ming~Y Lu, Drew~FK Williamson, Tiffany~Y Chen, Richard~J Chen, Matteo Barbieri, and Faisal Mahmood.
\newblock Data-efficient and weakly supervised computational pathology on whole-slide images.
\newblock \emph{Nature Biomedical Engineering}, 5\penalty0 (6):\penalty0 555--570, 2021.

\bibitem[Melendez et~al.(2015)Melendez, van Ginneken, Maduskar, Philipsen, Reither, Breuninger, Adetifa, Maane, Ayles, and S{\'{a}}nchez]{DBLP:journals/tmi/MelendezGMPRBAMAS15}
Jaime Melendez, Bram van Ginneken, Pragnya Maduskar, Rick H. H.~M. Philipsen, Klaus Reither, Marianne Breuninger, Ifedayo M.~O. Adetifa, Rahmatulai Maane, Helen Ayles, and Clara~I. S{\'{a}}nchez.
\newblock A novel multiple-instance learning-based approach to computer-aided detection of tuberculosis on chest x-rays.
\newblock \emph{{IEEE} Trans. Medical Imaging}, 34\penalty0 (1):\penalty0 179--192, 2015.
\newblock \doi{10.1109/TMI.2014.2350539}.

\bibitem[Morales et~al.(2021)Morales, Engan, and Naranjo]{DBLP:journals/dsp/MoralesEN21}
Sandra Morales, Kjersti Engan, and Valery Naranjo.
\newblock Artificial intelligence in computational pathology - challenges and future directions.
\newblock \emph{Digit. Signal Process.}, 119:\penalty0 103196, 2021.
\newblock \doi{10.1016/j.dsp.2021.103196}.

\bibitem[Quellec et~al.(2016)Quellec, Lamard, Cozic, Coatrieux, and Cazuguel]{DBLP:journals/tmi/QuellecLCCC16}
Gw{\'{e}}nol{\'{e}} Quellec, Mathieu Lamard, Michel Cozic, Gouenou Coatrieux, and Guy Cazuguel.
\newblock Multiple-instance learning for anomaly detection in digital mammography.
\newblock \emph{{IEEE} Trans. Medical Imaging}, 35\penalty0 (7):\penalty0 1604--1614, 2016.
\newblock \doi{10.1109/TMI.2016.2521442}.

\bibitem[Shao et~al.(2021)Shao, Bian, Chen, Wang, Zhang, Ji, and Zhang]{DBLP:conf/nips/ShaoBCWZJZ21}
Zhuchen Shao, Hao Bian, Yang Chen, Yifeng Wang, Jian Zhang, Xiangyang Ji, and Yongbing Zhang.
\newblock Transmil: Transformer based correlated multiple instance learning for whole slide image classification.
\newblock In Marc'Aurelio Ranzato, Alina Beygelzimer, Yann~N. Dauphin, Percy Liang, and Jennifer~Wortman Vaughan (eds.), \emph{Advances in Neural Information Processing Systems 34: Annual Conference on Neural Information Processing Systems 2021, NeurIPS 2021, December 6-14, 2021, virtual}, pp.\  2136--2147, 2021.

\bibitem[Shao et~al.(2023)Shao, Dai, Wang, Wang, and Zhang]{shao2023augdiff}
Zhuchen Shao, Liuxi Dai, Yifeng Wang, Haoqian Wang, and Yongbing Zhang.
\newblock Augdiff: Diffusion based feature augmentation for multiple instance learning in whole slide image, 2023.

\bibitem[Sharma et~al.(2021)Sharma, Shrivastava, Ehsan, Moskaluk, Syed, and Brown]{DBLP:conf/midl/SharmaSEMSB21}
Yash Sharma, Aman Shrivastava, Lubaina Ehsan, Christopher~A. Moskaluk, Sana Syed, and Donald~E. Brown.
\newblock Cluster-to-conquer: {A} framework for end-to-end multi-instance learning for whole slide image classification.
\newblock In Mattias~P. Heinrich, Qi~Dou, Marleen de~Bruijne, Jan Lellmann, Alexander Schlaefer, and Floris Ernst (eds.), \emph{Medical Imaging with Deep Learning, 7-9 July 2021, L{\"{u}}beck, Germany}, volume 143 of \emph{Proceedings of Machine Learning Research}, pp.\  682--698. {PMLR}, 2021.

\bibitem[Srinidhi et~al.(2019)Srinidhi, Ciga, and Martel]{DBLP:journals/corr/abs-1912-12378}
Chetan~L. Srinidhi, Ozan Ciga, and Anne~L. Martel.
\newblock Deep neural network models for computational histopathology: {A} survey.
\newblock \emph{CoRR}, abs/1912.12378, 2019.

\bibitem[Tellez et~al.(2021)Tellez, Litjens, van~der Laak, and Ciompi]{DBLP:journals/pami/TellezLLC21}
David Tellez, Geert Litjens, Jeroen van~der Laak, and Francesco Ciompi.
\newblock Neural image compression for gigapixel histopathology image analysis.
\newblock \emph{{IEEE} Trans. Pattern Anal. Mach. Intell.}, 43\penalty0 (2):\penalty0 567--578, 2021.
\newblock \doi{10.1109/TPAMI.2019.2936841}.

\bibitem[Tong et~al.(2014)Tong, Wolz, Gao, Guerrero, Hajnal, and Rueckert]{DBLP:journals/mia/TongWGGHR14}
Tong Tong, Robin Wolz, Qinquan Gao, Ricardo Guerrero, Joseph~V. Hajnal, and Daniel Rueckert.
\newblock Multiple instance learning for classification of dementia in brain {MRI}.
\newblock \emph{Medical Image Anal.}, 18\penalty0 (5):\penalty0 808--818, 2014.
\newblock \doi{10.1016/j.media.2014.04.006}.

\bibitem[Tourniaire et~al.(2023{\natexlab{a}})Tourniaire, Ilie, Hofman, Ayache, and Delingette]{DBLP:journals/mia/TourniaireIHAD23}
Paul Tourniaire, Marius Ilie, Paul Hofman, Nicholas Ayache, and Herv{\'{e}} Delingette.
\newblock {MS-CLAM:} mixed supervision for the classification and localization of tumors in whole slide images.
\newblock \emph{Medical Image Anal.}, 85:\penalty0 102763, 2023{\natexlab{a}}.
\newblock \doi{10.1016/j.media.2023.102763}.
\newblock URL \url{https://doi.org/10.1016/j.media.2023.102763}.

\bibitem[Tourniaire et~al.(2023{\natexlab{b}})Tourniaire, Ilie, Hofman, Ayache, and Delingette]{TOURNIAIRE2023102763}
Paul Tourniaire, Marius Ilie, Paul Hofman, Nicholas Ayache, and Hervé Delingette.
\newblock Ms-clam: Mixed supervision for the classification and localization of tumors in whole slide images.
\newblock \emph{Medical Image Analysis}, 85:\penalty0 102763, 2023{\natexlab{b}}.
\newblock ISSN 1361-8415.
\newblock \doi{https://doi.org/10.1016/j.media.2023.102763}.
\newblock URL \url{https://www.sciencedirect.com/science/article/pii/S1361841523000245}.

\bibitem[Tu et~al.(2019)Tu, Huang, He, and Zhou]{DBLP:journals/corr/abs-1906-04881}
Ming Tu, Jing Huang, Xiaodong He, and Bowen Zhou.
\newblock Multiple instance learning with graph neural networks.
\newblock \emph{CoRR}, abs/1906.04881, 2019.

\bibitem[Vaswani et~al.(2017)Vaswani, Shazeer, Parmar, Uszkoreit, Jones, Gomez, Kaiser, and Polosukhin]{DBLP:conf/nips/VaswaniSPUJGKP17}
Ashish Vaswani, Noam Shazeer, Niki Parmar, Jakob Uszkoreit, Llion Jones, Aidan~N. Gomez, Lukasz Kaiser, and Illia Polosukhin.
\newblock Attention is all you need.
\newblock In Isabelle Guyon, Ulrike von Luxburg, Samy Bengio, Hanna~M. Wallach, Rob Fergus, S.~V.~N. Vishwanathan, and Roman Garnett (eds.), \emph{Advances in Neural Information Processing Systems 30: Annual Conference on Neural Information Processing Systems 2017, December 4-9, 2017, Long Beach, CA, {USA}}, pp.\  5998--6008, 2017.

\bibitem[Wang et~al.(2021)Wang, Chen, Gao, Zhang, Guan, Dong, Zheng, Jiang, Yang, Wang, Huang, Ai, Yu, Li, Dong, Zhou, Liu, and Yu]{wang_predicting_2021}
Xiaodong Wang, Ying Chen, Yunshu Gao, Huiqing Zhang, Zehui Guan, Zhou Dong, Yuxuan Zheng, Jiarui Jiang, Haoqing Yang, Liming Wang, Xianming Huang, Lirong Ai, Wenlong Yu, Hongwei Li, Changsheng Dong, Zhou Zhou, Xiyang Liu, and Guanzhen Yu.
\newblock Predicting gastric cancer outcome from resected lymph node histopathology images using deep learning.
\newblock \emph{Nature Communications}, 12\penalty0 (1):\penalty0 1637, March 2021.
\newblock ISSN 2041-1723.
\newblock \doi{10.1038/s41467-021-21674-7}.
\newblock Number: 1 Publisher: Nature Publishing Group.

\bibitem[Wang et~al.(2018)Wang, Yan, Tang, Bai, and Liu]{DBLP:journals/pr/WangYTBL18}
Xinggang Wang, Yongluan Yan, Peng Tang, Xiang Bai, and Wenyu Liu.
\newblock Revisiting multiple instance neural networks.
\newblock \emph{Pattern Recognit.}, 74:\penalty0 15--24, 2018.
\newblock \doi{10.1016/j.patcog.2017.08.026}.

\bibitem[Xie et~al.(2020)Xie, Muhammad, Vanderbilt, Caso, Yarlagadda, Campanella, and Fuchs]{DBLP:conf/midl/XieMVCYCF20}
Chensu Xie, Hassan Muhammad, Chad~M. Vanderbilt, Raul Caso, Dig Vijay~Kumar Yarlagadda, Gabriele Campanella, and Thomas~J. Fuchs.
\newblock Beyond classification: Whole slide tissue histopathology analysis by end-to-end part learning.
\newblock In Tal Arbel, Ismail~Ben Ayed, Marleen de~Bruijne, Maxime Descoteaux, Herv{\'{e}} Lombaert, and Christopher Pal (eds.), \emph{International Conference on Medical Imaging with Deep Learning, {MIDL} 2020, 6-8 July 2020, Montr{\'{e}}al, QC, Canada}, volume 121 of \emph{Proceedings of Machine Learning Research}, pp.\  843--856, 2020.

\bibitem[Xiong et~al.(2021)Xiong, Zeng, Chakraborty, Tan, Fung, Li, and Singh]{DBLP:conf/aaai/XiongZCTFLS21}
Yunyang Xiong, Zhanpeng Zeng, Rudrasis Chakraborty, Mingxing Tan, Glenn Fung, Yin Li, and Vikas Singh.
\newblock Nystr{\"{o}}mformer: {A} nystr{\"{o}}m-based algorithm for approximating self-attention.
\newblock In \emph{Thirty-Fifth {AAAI} Conference on Artificial Intelligence, {AAAI} 2021, Thirty-Third Conference on Innovative Applications of Artificial Intelligence, {IAAI} 2021, The Eleventh Symposium on Educational Advances in Artificial Intelligence, {EAAI} 2021, Virtual Event, February 2-9, 2021}, pp.\  14138--14148. {AAAI} Press, 2021.
\newblock URL \url{https://ojs.aaai.org/index.php/AAAI/article/view/17664}.

\bibitem[Xu et~al.(2019)Xu, Song, Sun, Ku, Yang, Liu, Wang, Ma, and Xu]{DBLP:conf/iccv/XuSSKYLWMX19}
Gang Xu, Zhigang Song, Zhuo Sun, Calvin Ku, Zhe Yang, Cancheng Liu, Shuhao Wang, Jianpeng Ma, and Wei Xu.
\newblock {CAMEL:} {A} weakly supervised learning framework for histopathology image segmentation.
\newblock In \emph{2019 {IEEE/CVF} International Conference on Computer Vision, {ICCV} 2019, Seoul, Korea (South), October 27 - November 2, 2019}, pp.\  10681--10690, 2019.
\newblock \doi{10.1109/ICCV.2019.01078}.

\bibitem[Xu et~al.(2014)Xu, Zhu, Chang, Lai, and Tu]{DBLP:journals/mia/XuZCLT14}
Yan Xu, Jun{-}Yan Zhu, Eric~I{-}Chao Chang, Maode Lai, and Zhuowen Tu.
\newblock Weakly supervised histopathology cancer image segmentation and classification.
\newblock \emph{Medical Image Anal.}, 18\penalty0 (3):\penalty0 591--604, 2014.
\newblock \doi{10.1016/j.media.2014.01.010}.

\bibitem[Yao et~al.(2020)Yao, Zhu, Jonnagaddala, Hawkins, and Huang]{DBLP:journals/mia/YaoZJHH20}
Jiawen Yao, Xinliang Zhu, Jitendra Jonnagaddala, Nicholas~J. Hawkins, and Junzhou Huang.
\newblock Whole slide images based cancer survival prediction using attention guided deep multiple instance learning networks.
\newblock \emph{Medical Image Anal.}, 65:\penalty0 101789, 2020.
\newblock \doi{10.1016/j.media.2020.101789}.

\bibitem[Zhang et~al.(2022)Zhang, Meng, Zhao, Qiao, Yang, Coupland, and Zheng]{DBLP:conf/cvpr/ZhangMZQYCZ22}
Hongrun Zhang, Yanda Meng, Yitian Zhao, Yihong Qiao, Xiaoyun Yang, Sarah~E. Coupland, and Yalin Zheng.
\newblock {DTFD-MIL:} double-tier feature distillation multiple instance learning for histopathology whole slide image classification.
\newblock In \emph{{IEEE/CVF} Conference on Computer Vision and Pattern Recognition, {CVPR} 2022, New Orleans, LA, USA, June 18-24, 2022}, pp.\  18780--18790, 2022.
\newblock \doi{10.1109/CVPR52688.2022.01824}.

\bibitem[Zhang et~al.(2021)Zhang, Ma, Arnam, Gupta, Saltz, Vakalopoulou, and Samaras]{DBLP:conf/cvpr/ZhangMAGSVS21}
Jingwei Zhang, Ke~Ma, John S.~Van Arnam, Rajarsi Gupta, Joel~H. Saltz, Maria Vakalopoulou, and Dimitris Samaras.
\newblock A joint spatial and magnification based attention framework for large scale histopathology classification.
\newblock In \emph{{IEEE} Conference on Computer Vision and Pattern Recognition Workshops, {CVPR} Workshops 2021, virtual, June 19-25, 2021}, pp.\  3776--3784. Computer Vision Foundation / {IEEE}, 2021.
\newblock \doi{10.1109/CVPRW53098.2021.00418}.

\bibitem[Zhao et~al.()Zhao, Yang, Fang, Liu, Zhou, Zhang, Sun, Yang, Menze, Fan, and Yao]{DBLP:conf/cvpr/ZhaoYFLZZSYMFY20}
Yu~Zhao, Fan Yang, Yuqi Fang, Hailing Liu, Niyun Zhou, Jun Zhang, Jiarui Sun, Sen Yang, Bjoern~H. Menze, Xinjuan Fan, and Jianhua Yao.
\newblock Predicting lymph node metastasis using histopathological images based on multiple instance learning with deep graph convolution.
\newblock In \emph{2020 {IEEE/CVF} Conference on Computer Vision and Pattern Recognition, {CVPR} 2020, Seattle, WA, USA, June 13-19, 2020}, pp.\  4836--4845. Computer Vision Foundation / {IEEE}.
\newblock \doi{10.1109/CVPR42600.2020.00489}.

\bibitem[Zheng et~al.(2022)Zheng, Gindra, Green, Burks, Betke, Beane, and Kolachalama]{DBLP:journals/tmi/ZhengGGBBBK22}
Yi~Zheng, Rushin~H. Gindra, Emily~J. Green, Eric~J. Burks, Margrit Betke, Jennifer~E. Beane, and Vijaya~B. Kolachalama.
\newblock A graph-transformer for whole slide image classification.
\newblock \emph{{IEEE} Trans. Medical Imaging}, 41\penalty0 (11):\penalty0 3003--3015, 2022.
\newblock \doi{10.1109/TMI.2022.3176598}.
\newblock URL \url{https://doi.org/10.1109/TMI.2022.3176598}.

\bibitem[Zhou et~al.(2021)Zhou, Jin, Chen, Huang, Huang, Wang, Zhao, Chen, Guo, and Liao]{DBLP:journals/cmig/ZhouJCHHWZCGL21}
Changjiang Zhou, Yi~Jin, Yuzong Chen, Shan Huang, Rengpeng Huang, Yuhong Wang, Youcai Zhao, Yao Chen, Lingchuan Guo, and Jun Liao.
\newblock Histopathology classification and localization of colorectal cancer using global labels by weakly supervised deep learning.
\newblock \emph{Comput. Medical Imaging Graph.}, 88:\penalty0 101861, 2021.
\newblock \doi{10.1016/j.compmedimag.2021.101861}.

\end{thebibliography}
